%% file: main.tex
\documentclass[10pt,twocolumn,letterpaper]{article}

\usepackage[pagenumbers]{cvpr} 

\definecolor{cvprblue}{rgb}{0.21,0.49,0.74}
\usepackage[pagebackref,breaklinks,colorlinks,citecolor=cvprblue]{hyperref}
\usepackage{capt-of}
\usepackage[dvipsnames]{xcolor}
\usepackage{xspace}
\usepackage{enumitem}
\usepackage{amsmath,amssymb,amsbsy,amsfonts,dsfont,pifont,bm,bbm,mathrsfs,mathtools,nicefrac}
\usepackage{algorithm,algpseudocode,listings}
\usepackage{booktabs,multirow,adjustbox,diagbox,threeparttable,tabularray,colortbl}

\definecolor{Gray}{gray}{0.94}
\definecolor{liGray}{gray}{0.5}
\definecolor{LightCyan}{rgb}{0.88,1,1}

\newlength\savewidth

\usepackage{indentfirst} 
\newcommand{\method}{\texttt{UniAnimate-DiT}\xspace}

\title{UniAnimate-DiT: Human Image Animation with Large-Scale Video Diffusion Transformer}

\author{
   \hspace{-0.4cm} 
   Xiang Wang$^{1}$
     \hspace{0.01cm} 
    Shiwei Zhang$^{2}$
     \hspace{0.01cm} 
    Longxiang Tang$^{3}$
     \hspace{0.01cm} 
    Yingya Zhang$^2$ 
     \hspace{0.01cm}
    Changxin Gao$^1$ \\
    \hspace{-0.4cm} 
    Yuehuan Wang$^1$ 
     \hspace{0.01cm}
     Nong Sang$^{1}$
      \vspace{2mm}
     \\
    $^1$Key Laboratory of Image Processing and Intelligent Control,\\   School of Artificial Intelligence and Automation, Huazhong University of Science and Technology\\
      $^2$Alibaba Group \hspace{0.5cm} $^3$Tsinghua University \\
}

\begin{document}

\twocolumn[{
\maketitle
    \centering
    \vspace{-12pt}
    \includegraphics[width=0.99\textwidth]{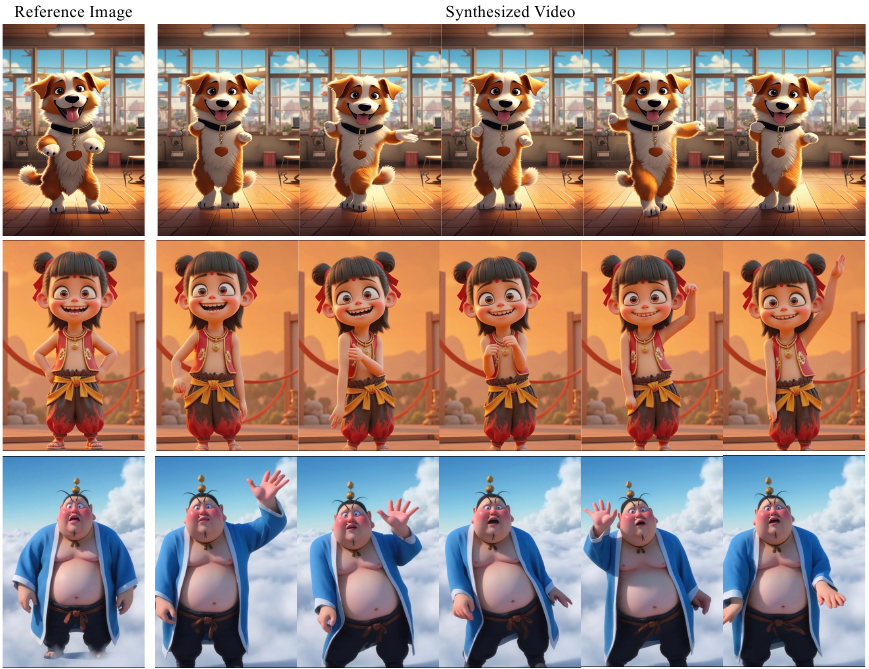}
    \vspace{-7pt}
    \captionof{figure}{
        \textbf{Image animation examples} synthesized by the proposed \method with Wan2.1-I2V-14B~\cite{wang2025wanvideo} as the base model.
    }
    \label{first_figure}
    \vspace{15pt}
    }
]

\input{sec/0_abstract}    
\input{sec/1_intro}

\input{sec/2_method}

\input{sec/3_experiment}

\input{sec/4_conclusion}
{
    \small
    \bibliographystyle{ieeenat_fullname}
    \bibliography{main}
}


\end{document}

%% file: sec/0_abstract.tex
\begin{abstract}

\begin{figure*}[th]
    \centering
    \includegraphics[width=0.99\linewidth]{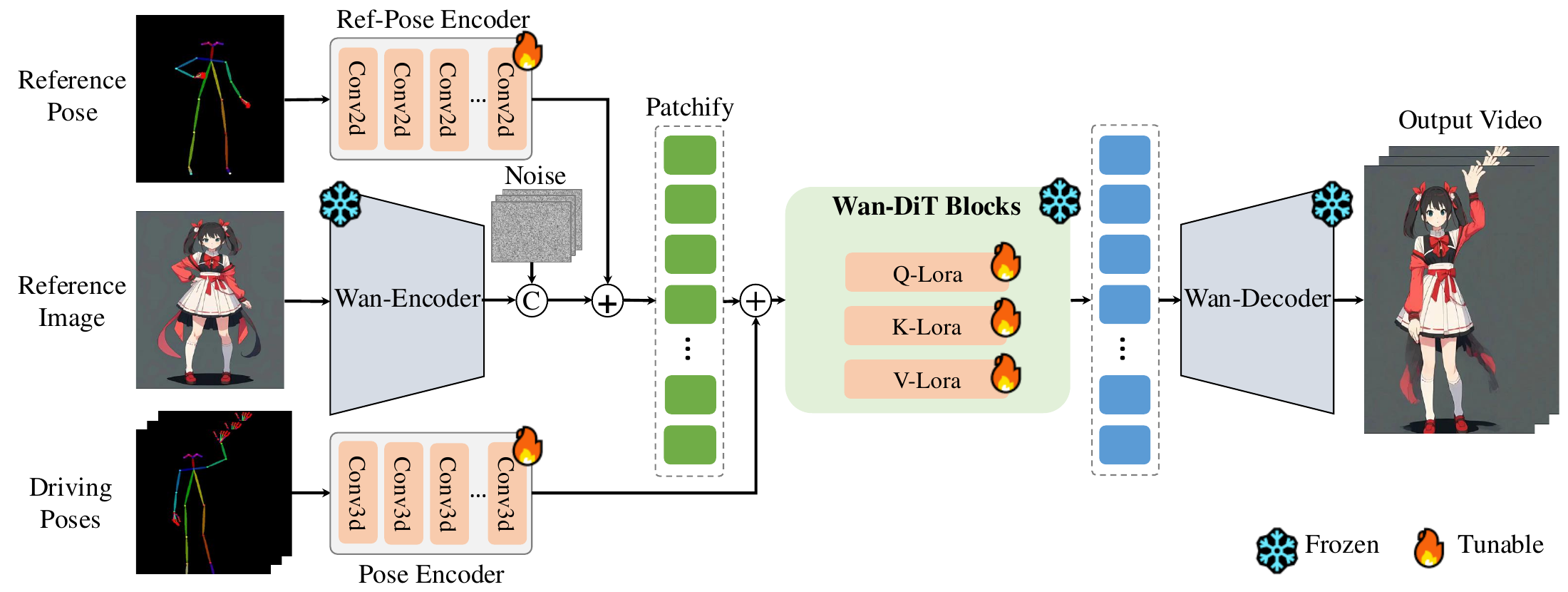}
    \caption{The overall architecture of the proposed \method base on Wan2.1 model.}
    \label{fig:network}
\end{figure*}

This report presents \method, an advanced project that leverages the cutting-edge and powerful capabilities of the open-source Wan2.1 model for consistent human image animation.
%
Specifically, to preserve
the robust generative capabilities of the original Wan2.1 model,
we implement Low-Rank Adaptation (LoRA) technique to fine-tune a
minimal set of parameters, significantly reducing training memory overhead.
A lightweight pose encoder consisting of multiple stacked 3D convolutional layers is designed to encode motion information of driving poses. 
Furthermore, we adopt a simple concatenation operation to integrate the reference appearance into the model and incorporate the pose information of the reference image for enhanced pose alignment. 
Experimental results show that our approach achieves visually appearing and temporally consistent high-fidelity animations. 
Trained on 480p (832x480) videos, \method demonstrates strong generalization capabilities to seamlessly upscale to 720P (1280x720) during inference.
%
The training and inference code is publicly available at \url{https://github.com/ali-vilab/UniAnimate-DiT}.

\end{abstract}

%% file: sec/1_intro.tex
\section{Introduction}
\label{sec:intro}

Human image animation has undergone significant advances with the convergence of generative modeling techniques~\cite{yang2018pose,lin2025exploring,peebles2023scalable}, especially with the rise of diffusion models~\cite{karras2023dreampose,unianimate,Animateanyone,xu2024magicanimate,zhang2023i2vgen,kong2024hunyuanvideo,imagenvideo,champ,VideoLDM}. 
This task aims to enable the transformation of a static reference image into dynamic video sequences that depict lifelike, temporally consistent movements that adhere to the guidance of driving poses. 

Traditional methods~\cite{Animateanyone,unianimate,xu2024magicanimate,tan2024animate} in this domain often leverage a 3D-UNet base model to generate videos, struggling with temporal coherence and realism.
For example, UniAnimate~\cite{unianimate} presents a unified framework based on a 3D-UNet TF-T2V~\cite{tft2v} model to encode reference appearance and motion movement.
The animation performance may be constrained by the capability of the base model.
This motivates a shift towards more advanced video generative models.
The transition to a more advanced Diffusion Transformer (DIT)-based model, such as Wan2.1~\cite{wang2025wanvideo}, provides a potential direction to enhance the quality of generated videos. 

To this end,
this report aims to fulfill this gap and presents \method, a simple but effective framework based on Wan2.1 for consistent human image animation.
Specifically, we employ LoRAs to finetune a smaller set of model parameters, reducing training memory overhead while maintaining the original model's generative potency.
A lightweight pose encoder consisting of multiple stacked 3D convolutional layers is used to encode driving motion information. 
In addition,
the reference pose information is also incorporated to enhance appearance alignment.
Qualitative experimental results (~\cref{first_figure}) show that our
approach achieves visually appearing and temporally consistent high-fidelity animations. 
Despite being trained at
480P (832x480) video resolution, \method has
the ability to seamlessly upscale to 720P (1280x720) during inference.

%% file: sec/2_method.tex
\section{Method}

\begin{figure*}[th]
    \centering
    \includegraphics[width=0.99\linewidth]{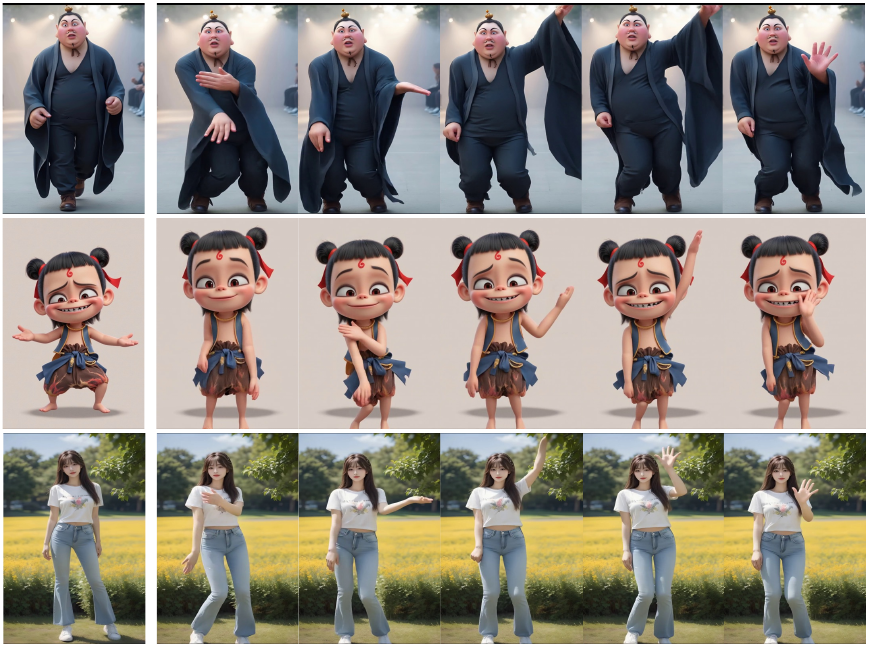}
    \vspace{-3mm}
    \caption{Video cases synthesized by the proposed \method.}
    \label{fig:figure}
\end{figure*}

\begin{figure*}[th]
    \centering
    \includegraphics[width=0.99\linewidth]{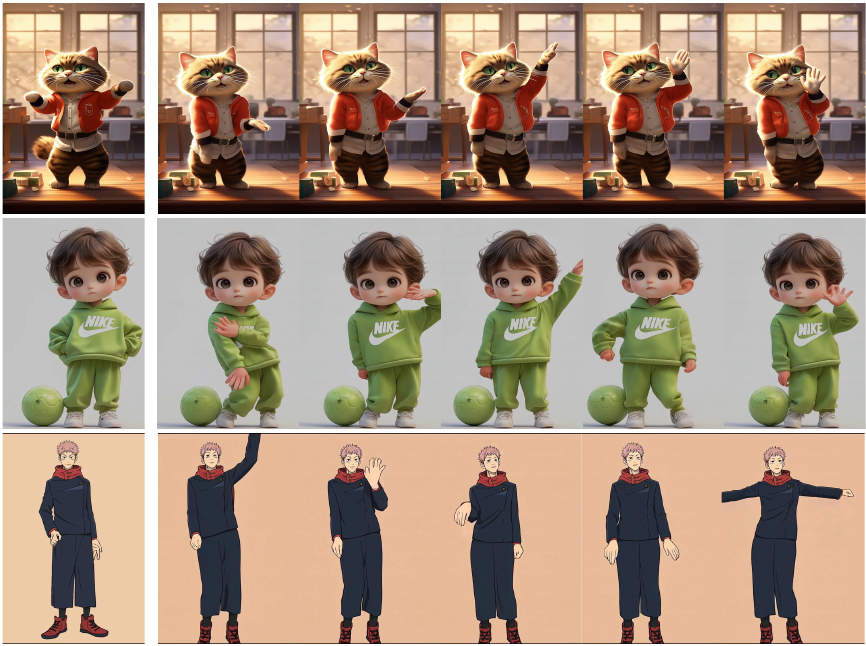}
    \vspace{-3mm}
    \caption{Video cases synthesized by the proposed \method.}
    \label{fig:figure2}
\end{figure*}

The overall architecture of \method is displayed in~\cref{fig:network},
which is designed to adapt the human image animation task by incorporating the advanced video DiT model.
The framework comprises several key components:
1) Video Diffusion Transformer Model (Wan2.1): The video DiT model serves as the primary generative engine, providing a robust mechanism for high-quality output generation.
2) LoRA Fine-tuning: Low-Rank Adaptation (LoRA) is an efficient parameter tuning technique that reduces the number of trainable parameters. Through LoRA, we fine-tune a limited number of parameters, reducing memory overhead and enhancing adaptability without sacrificing performance.
3) Pose Encoder: This lightweight module consists of multiple stacked 3D convolutional layers designed to extract temporal and spatial features of driving poses effectively.
4) Ref-Pose Encoder: the reference pose information is also incorporated by stacked 2D convolutional layers to enhance appearance alignment. Reference pose information is incorporated by summing it with a noisy latent vector.

\vspace{1mm}
\noindent \textbf{Discussions}. The pose encoder is a critical component that enables accurate representation of human poses and movement dynamics. In our setting, the encoder consists of seven layers of 3D convolutions. Through experiments, we observe that a lower number of layers (e.g., four layers) resulted in a limited receptive field, which hinders the model’s ability to control the generated animation accurately. Thus, a deeper architecture improves the model's understanding of temporal contexts and enhances motion control.
Initially, the driven pose features were concatenated directly with the noisy latent vector (which is 16-dimensional), resulting in ineffective control over the generated animations due to its limited feature representation capacity. Recognizing this limitation, we try to inject reference pose information into the model at a more meaningful feature level. Finally,  we choose to integrate pose information into the patchified tokens (5120-dimension). This adjustment provides a significantly richer representation, improving the model's ability to learn and control detailed pose characteristics during the generation phase.


\vspace{1mm}
\noindent \textbf{Long video generation}. Our \method also supports long video generation by applying the overlapped slide window
strategy~\cite{bar2023multidiffusion}.
The first frame feature after the Wan-VAE only represents one frame.
To improve the consistency between each window, the first two frame features of the subsequent windows are discarded.


%% file: sec/3_experiment.tex
\section{Experiments}

\subsection{Dataset and setup}
We collect a video dataset that contains about 10K human dance videos to train our \method.
The dataset used for training consists of diverse human images annotated with key poses
and spans a wide range of actions and lighting conditions, allowing the model to learn a comprehensive set of animation dynamics. 
In our experiments, 6 Nvidia A800 GPUs are leveraged. 
The model is trained at 832x480 resolution due to GPU memory restriction.
Please refer to our open source code for more  training and inference details.

\subsection{Qualitative evaluation}
An important feature of our framework is its capability to generalize to large resolutions during inference. While training occurs at 480P, our model can upscale outputs to 720P effortlessly. This scalability broadens the practical applications of our technology for higher-resolution video generation.
In~\cref{fig:figure,fig:figure2}, we show the qualitative evaluations of our method.
The
generated videos reveal a high degree of fidelity and continuity, demonstrating lifelike movements.
These results indicate the effectiveness of the proposed \method.


%% file: sec/4_conclusion.tex
\section{Conclusion}

\method establishes a significant advancement in human image animation by effectively combining the capabilities of large-scale video DiTs with efficient fine-tuning techniques. Our open-source implementation provides a valuable resource for researchers and developers in the field, further enabling the evolution of realistic animation technologies. 

\section*{Acknowledgment}
This work is supported by the National Natural Science Foundation
of China under grants U22B2053 and 623B2039, and Alibaba Group through Alibaba Research Intern Program.